\def\year{2021}\relax
\documentclass[letterpaper]{article} 
\usepackage{aaai21}  
\usepackage{times}  
\usepackage{courier}  
\usepackage[hyphens]{url}  
\usepackage{graphicx} 
\usepackage{natbib}  
\usepackage{caption} 
\usepackage{CJK}
\usepackage{makecell}
\newcommand{\tabincell}[2]{\begin{tabular}{@{}#1@{}}#2\end{tabular}}
\title{NaturalConv: A Chinese Dialogue Dataset Towards \\ Multi-turn Topic-driven Conversation}

\author{
    Xiaoyang Wang\footnote{Xiaoyang Wang and Chen Li contributed equally to this work.},
    Chen Li\footnotemark[1],
    Jianqiao Zhao,
    Dong Yu
    \\
}
\affiliations{
    Tencent AI Lab, Bellevue, WA\\
    \{shawnxywang, ailabchenli, markjzhao, dyu\}@tencent.com
}

\begin{document}
\maketitle
\begin{CJK*}{UTF8}{gbsn}

\begin{abstract}
In this paper, we propose a Chinese multi-turn topic-driven conversation dataset, NaturalConv, which allows the participants to chat anything they want as long as any element from the topic is mentioned and the topic shift is smooth. Our corpus contains 19.9K conversations from six domains, and 400K utterances with an average turn number of 20.1. These conversations contain in-depth discussions on related topics or widely natural transition between multiple topics. We believe either way is normal for human conversation. To facilitate the research on this corpus, we provide results of several benchmark models. Comparative results show that for this dataset, our current models are not able to provide significant improvement by introducing background knowledge/topic. Therefore, the proposed dataset should be a good benchmark for further research to evaluate the validity and naturalness of multi-turn conversation systems. Our dataset is available\footnote{Also at \url{https://huggingface.co/datasets/xywang1/NaturalConv}} at \url{https://ailab.tencent.com/ailab/nlp/dialogue/#datasets}.
\end{abstract}

\section{Introduction}
There is a resurgent interest in developing open domain dialogue system,  due to the availability of large amounts of conversational data and the recent progress on neural approaches \cite{huang2019challenges}.  
However, building open-domain dialogue systems that can converse on various topics like humans remains extremely challenging and most current open domain dialogue system are only good at generating general responses without too much meaningful information~\cite{gao2019neural}. 

Therefore, increasing research efforts are moving to consider how to incorporate various kinds of information to improve the quality of open domain conversation. The information includes but not limited to personality \cite{qian2018assigning}, common sense \cite{zhou2018commonsense}, reasoning \cite{zhou2018interpretable}, emotion\cite{zhou2018emotional}, extra knowledge in terms of knowledge graph \cite{moon-etal-2019-opendialkg, wu-etal-2019-proactive, zhou2020kdconv} etc. Especially,  a variety of knowledge grounded dialogue corpora \cite{zhu2017flexible, dinan2019wizard, liu-etal-2018-knowledge, moghe-etal-2018-towards, zhou-etal-2018-dataset,   moon-etal-2019-opendialkg, qin2019conversing, tuan-etal-2019-dykgchat, wu-etal-2019-proactive, zhou2020kdconv}  have been released to demonstrate attempts for generating informative responses with topic related dialogue. Knowledge/Topic grounded conversation is a kind of new type conversation. On one hand, unlike open domain dialogue, it involves some specific topics which need extra knowledge during response generation. On the other hand, it also has various indirect information to the topic such as chitchat, saying a joke, expressing personal experiences, etc. 

Therefore, we believe this kind of topic grounded conversations are closer to human-like conversations than open-domain dialogue in terms of naturalness and popularity. However, we find two common drawbacks from current available grounded conversation corpora: one is that almost all the mentioned work requires that the participants can only chat within the given topic and assumes the participants are familiar with the topic by reading the provided document or knowledge graph. But in our real life, people can easily extend the topic if they are very familiar with that and can also easily shift to other topics if their partner starts an unfamiliar topic. The other shortcoming is that most mentioned previous work encourages annotators to directly talk about the topic once they start the conversation. The work~\cite{moghe-etal-2018-towards} even explicitly forbids chitchat during the annotation stage. However, the reality is quite opposite. Few daily conversations between people will directly step into topic after start. People will always have greeting before any formal or informal talk.

In order to conquer these problems, we propose NaturalConv, a Chinese dialogue dataset towards multi-turn topic-driven conversation with scenario. It is quite suitable for modeling topic interactions in multi-turn natural human-like dialogues. The common point with previous corpora is that the proposed dataset is also topic grounded and collected by annotators who communicate based on one given topic in the form of a news article. But the biggest different characteristics are the following: First, talk does not have to be only about the content from the article if one or both of them are not interested in that topic. Participants can talk about anything they want as long as any information from the news article is mentioned and the transition among the topic is natural;
Second, we require that two participants need to suppose a scenario for their conversation.
It means annotators conduct the dialogue task like a cosplay. They can assume the talk happens in any scenario as long as it follows a normal logic to conduct the conversation; Third, we allow chitchat/greetings in our conversation.

Table~\ref{oneexample} shows an example. The top is the news article which both participants can access. From the dialogue, we can guess this dialogue happens between two students before the first class in the morning. Second, we find only 2 (A-4, A-5) out of 20 utterances explicitly mention the specific information of the news. Other utterances include chitchat (A-1, B-1), query/response about scenario (A-9, B-9, A-10, B-10), and personal experience related to topic (A-7, A-8, B-8), etc. But we can observe that the whole dialogue is more natural and human-like than most previous multi-turn dialogues that are more QA-like information exchange.

\begin{table*}[!htbp]
  \centering
  \small
  \begin{tabular}{cp{1.6\columnwidth}p{0.4\columnwidth}}
  \hline
  \multicolumn{3}{p{2.1\columnwidth}}{北京时间今天凌晨，荷甲第4轮进行了两场补赛。阿贾克斯和埃因霍温均在主场取得了胜利，两队7轮后同积17分，阿贾克斯以6个净胜球的优势领跑积分榜。
0点30分，埃因霍温与格罗宁根的比赛开战，埃因霍温最终3-1主场获胜。2点45分，阿贾克斯主场与福图纳锡塔德之战开球。由于埃因霍温已经先获胜了，阿贾克斯必须获胜才能在积分榜上咬住对方。
在整个上半场，阿贾克斯得势不得分，双方0-0互交白卷。在下半场中，阿贾克斯突然迎来了大爆发。在短短33分钟内，阿贾克斯疯狂打进5球，平均每6分钟就能取得1个进球。
在第50分钟时，新援普罗梅斯为阿贾克斯打破僵局。塔迪奇左侧送出横传，普罗梅斯后点推射破门。53分钟亨特拉尔头球补射，内雷斯在门线前头球接力破门。
68分钟时，普罗梅斯近距离补射梅开二度。这名27岁的前场多面手，跑到场边来了一番尬舞。77分钟时阿贾克斯收获第4球，客队后卫哈里斯在防传中时伸腿将球一捅，结果皮球恰好越过门将滚入网窝。
在第83分钟时，普罗梅斯上演了帽子戏法，比分也最终被定格为5-0。在接到塔迪奇直传后，普罗梅斯禁区左侧反越位成功，他的单刀低射从门将裆下入网。普罗梅斯这次的庆祝动作是秀出三根手指，不过他手指从上到下抹过面部时的动作，很有点像是在擦鼻涕。In the early morning of Beijing time, the Dutch league had two matches in the fourth round.  Ajax and Eindhoven both won at home, scoring 17 points after 7 rounds. Ajax led the scoreboard with 6 goals advantage.
At 0:30, Eindhoven played with Groningen, and Eindhoven won 3-1 at home.  At 2:45, Ajax kicked off the match against Fortuna's Sitad at home.  Since Eindhoven had won first, Ajax must win in order to keep the pace.
In the first half, neither team scored.  In the second half, Ajax started to score goals.  In a short span of 33 minutes, Ajax scored five goals crazily, averaging one goal every 6 minutes.
In the 50th minute, Promes broke the deadlock for Ajax.  Tadic sent out a cross on the left, and Promes pushed back and scored.  Huntelaar made a header shot in the 53rd minute, and Neres headed the ball into the net in front of the goal line.
In the 68th minute, Promes scored for a second time.  The 27-year-old versatile frontcourt player ran to the edge of the field for a full dance.  Ajax scored the fourth goal in the 77th minute. Away team defender Harris gave the ball a poke with his leg while defending the cross. As a result, the ball just passed the goalkeeper and rolled into the net.
In the 83rd minute, Promes achieved a hat-trick and the score was finally fixed at 5-0.  After receiving Tadic's direct pass, Promes successfully countered offside on the left side of the penalty area. His one-on-one low shot went into the net from under the goalkeeper's crotch. Promes showed three fingers for celebration, but his fingers rubbed his face from top to bottom, which was a bit like wiping his nose.} \\ \hline
  \multicolumn{1}{|c|}{Turn}&\multicolumn{1}{c|}{Content of Dialogue} &\multicolumn{1}{c|}{Description}\\ \hline
  \multicolumn{1}{|c|}{A-1}&\multicolumn{1}{p{1.2\columnwidth}|}{嗨，你来的挺早啊。(Hi, you come so early.)} &
  \multicolumn{1}{p{0.6\columnwidth}|}{chitchat} \\\hline
      \multicolumn{1}{|c|}{B-1}&\multicolumn{1}{p{1.2\columnwidth}|}{是啊，你怎么来得这么晚？(Yes, why do you come so late?)}  & \multicolumn{1}{p{0.6\columnwidth}|}{chitchat}\\\hline
    \multicolumn{1}{|c|}{A-2}&\multicolumn{1}{p{1.2\columnwidth}|}{昨晚我看了球赛，所以今早起晚了，也没吃饭。(I watched a sports game last night, so I woke up late this morning, and I have not had my breakfast.)}  & \multicolumn{1}{p{0.6\columnwidth}|}{chitchat, start to introduce the topic about soccer}\\\hline
    \multicolumn{1}{|c|}{B-2}&\multicolumn{1}{p{1.2\columnwidth}|}{现在这个点食堂应该有饭，你看可什么球赛啊？篮球吗？(Oh, the cafeteria should still be open now. Which game did you watch? Basketball game?)}  &\multicolumn{1}{p{0.6\columnwidth}|}{}\\\hline
    \multicolumn{1}{|c|}{A-3}&\multicolumn{1}{p{1.2\columnwidth}|}{不是，足球。(No, soccer game.)} &\multicolumn{1}{p{0.6\columnwidth}|}{introduce general information of news} \\\hline
    \multicolumn{1}{|c|}{B-3}&\multicolumn{1}{p{1.2\columnwidth}|}{怪不得，足球时间长。(I see. Soccer game usually takes longer than basketball game.)}&\multicolumn{1}{p{0.6\columnwidth}|}{} \\\hline
    \multicolumn{1}{|c|}{A-4}&\multicolumn{1}{p{1.2\columnwidth}|}{你知道么，每次都是普罗梅斯进球。(Do you know, Promes can score every time.)} &\multicolumn{1}{p{0.6\columnwidth}|}{provide a specific information} \\\hline
    \multicolumn{1}{|c|}{B-4}&\multicolumn{1}{p{1.2\columnwidth}|}{这个我刚才也看了新闻了，他好有实力啊。(Oh, I also read the news of that game. Yes. He is very strong.)}&\multicolumn{1}{p{0.6\columnwidth}|}{} \\\hline
    \multicolumn{1}{|c|}{A-5}&\multicolumn{1}{p{1.2\columnwidth}|}{是啊，尤其是他那个帽子戏法，让我看的太惊心动魄了。(Yes, especially he had a hat-trick last night. I was so excited.)} &\multicolumn{1}{p{0.6\columnwidth}|}{provide a specific information} \\\hline
    \multicolumn{1}{|c|}{B-5}&\multicolumn{1}{p{1.2\columnwidth}|}{我一同学在群里说了，每次聊天都离不开他，可见他的实力有多强大。(One of my classmates always talked about Promes every time as well, which means how strong he is.)
}  &\multicolumn{1}{p{0.6\columnwidth}|}{the key point of this utterance is not related to news} \\\hline
\multicolumn{1}{|c|}{A-6}&\multicolumn{1}{p{1.2\columnwidth}|}{是啊，看来你那个同学和我是一样的想法。(Yes, your classmate has the same point with me.)}  &\multicolumn{1}{p{0.6\columnwidth}|}{from this utterance to the end, none is related to the news, but still about the soccer topic.} \\\hline
    \multicolumn{1}{|c|}{B-6}&\multicolumn{1}{p{1.2\columnwidth}|}{我好不容易摆脱他的话题，你又来一个说出他的名字。(I had a hard time getting rid of his topic, and now you mentioned his name again.)
}  &\multicolumn{1}{p{0.6\columnwidth}|}{
} \\\hline

    \multicolumn{1}{|c|}{A-7}&\multicolumn{1}{p{1.2\columnwidth}|}{哈哈，你不懂我们对足球有多热爱。(Haha, you don't understand how much we love soccer.)
}  &\multicolumn{1}{p{0.6\columnwidth}|}{} \\\hline

    \multicolumn{1}{|c|}{B-7}&\multicolumn{1}{p{1.2\columnwidth}|}{我知道你热爱，我还记得你参加初中比赛还拿到冠军呢。你功不可没啊。(I know you love it, and I still remember you won the junior high school competition. You played a very important role.)
}  &\multicolumn{1}{p{0.6\columnwidth}|}{} \\\hline
    \multicolumn{1}{|c|}{A-8}&\multicolumn{1}{p{1.2\columnwidth}|}{哈哈，还是你能记得我当时的辉煌。(Haha, you can still remember my glory at that time.)
} &\multicolumn{1}{p{0.6\columnwidth}|}{
} \\\hline
    \multicolumn{1}{|c|}{B-8}&\multicolumn{1}{p{1.2\columnwidth}|}{没办法，咱俩从小一起长大的，彼此太了解彼此了。(We grew up together, and we know each other too well.)
} &\multicolumn{1}{p{0.6\columnwidth}|}{
} \\\hline
    \multicolumn{1}{|c|}{A-9}&\multicolumn{1}{p{1.2\columnwidth}|}{嗯，老师来了。(Sure. The professor is coming.)
} &\multicolumn{1}{p{0.6\columnwidth}|}{
} \\\hline

    \multicolumn{1}{|c|}{B-9}&\multicolumn{1}{p{1.2\columnwidth}|}{快打开课本，老师要检查。(Open the book. He said he would check our work this time.)}&\multicolumn{1}{p{0.6\columnwidth}|}{}   \\\hline
    \multicolumn{1}{|c|}{A-10}&\multicolumn{1}{p{1.2\columnwidth}|}{嗯嗯，下课再聊。(OK. Let's talk after class.)} &\multicolumn{1}{p{0.6\columnwidth}|}{} \\\hline
    \multicolumn{1}{|c|}{B-10}&\multicolumn{1}{p{1.2\columnwidth}|}{嗯。(Sure.)} &\multicolumn{1}{p{0.6\columnwidth}|}{} \\\hline
\end{tabular}
\caption{One example from our NaturalConv dataset.}
\label{oneexample}
\end{table*}

With above mentioned significantly different properties, we also find it is not trivial to incorporate the document/knowledge into the process of dialogue generation. Our implemented methods of incorporating document/knowledge discussed in the Methods part of the paper do not bring in significant performance gains on our dataset. In summary, this paper makes the following contributions:

\begin{itemize}
\item We collect a new dataset, NaturalConv, based on topic-driven conversation generation in Chinese. It is much closer to human-like conversation with natural property including a full and natural setting such as scenario assumption, free topic extension, greetings, ect. It contains about 400K utterances and 19.9K dialogues in multiple domains (including but not limited to sports, entertainment, and technology). The averaged turn number is 20, remarkably longer than those in other corpora.
\item NaturalConv provides a benchmark to evaluate the ability of generating conversations in a natural setting.  The corpus can empower the future research of not only document grounded conversation generation, but also conversation style and strategy learning from different scenarios.
\item We also conduct extensive experiments on this corpus to facilitate future research. Results show it is still very challenging to incorporate document knowledge on our dataset for dialogue generation. We still need deep research work to improve the system to handle such natural and vivid conversation.
\end{itemize}

\section{Related Work}
As more dialogue data is accessible and huge computation resource is available, neural based open-domain conversation generation has been largely advanced recently \cite{adiwardana2020humanlike, roller2020recipes}. But, most neural response generation models developed for open-domain dialogue systems are not grounded in real world, which prevents these systems from effectively conversing about anything meaningful. Knowledge grounding is crucial for the system to provide practical responses. Otherwise, the system would prefer bland and repetitive responses.

To accelerate the research of knowledge-grounded conversation, several knowledge-grounded corpora are proposed. Some \cite{ghazvininejad2018knowledge, liu-etal-2018-knowledge, tuan-etal-2019-dykgchat, qin2019conversing} obtain the knowledge by automatic methods such as NER and string match. But more \cite{zhou-etal-2018-dataset, dinan2019wizard, gopalakrishnan2019topical, moon-etal-2019-opendialkg, wu-etal-2019-proactive,zhou2020kdconv} collect knowledge during the annotation from annotators.

There are also differences among these corpus. From the aspect of whether two participants or only one participant can access the knowledge, \cite{dinan2019wizard} assumes one annotator is a professional person who can access Wikipedia resource, while the other one is an apprentice who is seeking information and knows nothing about the topic. On the other hand, \cite{moon-etal-2019-opendialkg, wu-etal-2019-proactive, zhou2020kdconv} allow all annotators to access knowledge.

The knowledge in \cite{zhou-etal-2018-dataset, dinan2019wizard, gopalakrishnan2019topical} is unstructured plain text, while \cite{moon-etal-2019-opendialkg, wu-etal-2019-proactive, zhou2020kdconv} provides structured knowledge graph. \cite{moon-etal-2019-opendialkg} uses Freebase~\cite{bast2014easy} as background knowledge. To the best of our knowledge, DuConv \cite{wu-etal-2019-proactive} and KdConv \cite{zhou2020kdconv} are the only two existing Chinese human-labeled knowledge-grounded dialogue datasets. The DuConv utilizes the combination of unstructured text like short comments and structured knowledge graphs as knowledge resources. One limitation of DuConv is its strong assumption that the conversation must transfer from one entity to another inside the knowledge graph, while this is not always true during human conversation. KdConv constructs their knowledge graph from multiple resources. One defect of KdConv is its high overlap between dialogue and the provided knowledge, which means the annotator heavily duplicates the content from knowledge graph and the dialogue lacks variability. Table~\ref{corpus_cmp} shows statistics among corpora that share similar settings with ours.

\begin{table*}[ht]
\begin{center}
\begin{tabular}{ccccccc} \hline
 \textbf{Dataset}  &  \textbf{Language} & \tabincell{c}{ \textbf{Document} \\  \textbf{Type}} & \tabincell{c}{ \textbf{Annotation}\\  \textbf{Level}} &  \textbf{Topic} &  \textbf{Avg. \# turns} & \ \textbf{\# uttrs} \\ \hline
CMU DoG & English   & Text & Sentence & Film & 22.6 & 130k  \\
Wizard of Wiki & English   & Text & Sentence & Multiple & 9.0 & 202k\\
DuConv & Chinese   & Text\&KG & Dialogue &  Film & 5.8 & 91k\\
KdConv & Chinese   & Text\&KG & Sentence &  \tabincell{c}{Film, Music,\\ Travel}& 19.0 & 86k\\ \hline
 \textbf{NaturalConv} &  \textbf{Chinese}  &  \textbf{Text} &  \textbf{Dialogue} &  \tabincell{c}{ \textbf{Sports, Ent,}\\  \textbf{Tech, Games,} \\  \textbf{Edu, Health}} &  \textbf{20.1} &  \textbf{400k}\\ \hline
\end{tabular}
\caption{Comparison among our NaturalConv corpus and other human-labeled document/knowledge grounded dialogue corpus.}
\label{corpus_cmp}
\end{center}
\end{table*}

\section{Dataset}
In this section, we describe the creation of NaturalConv in details. NaturalConv is designed to collect a multi-turn document grounded dialogue dataset with scenario and naturalness property of conversation. The created dialogues are expected to include three key points: dialogue with meaningful content, dialogue in a scenario, and dialogue with naturalness. In the following, we will describe how we design the data collection.

\subsection{Dialogue Collection}
\textbf{Collect and filter document}: First of all, we believe only with a common topic, the dialogue can have meaningful content. Therefore, we collect news articles as the grounding documents for dialogue. At the same time, we avoid to choose professional materials or topics because we believe most daily conversations are leisure talk about what happens daily. As a result, we collect in total of 6,500 news articles that are in 6 categories in time range from September 2019 to December 2019. Table~\ref{corpus_sta} shows the distribution of each category. The uneven distribution is caused by the popularity of different categories and the appropriateness of news articles for dialogue grounding article. For example, we filter politics and economics news due to their sensibility, remove short news because of their poor informativeness, and drop too long news to avoid annotators spending too much time on reading,  etc.

\textbf{Create dialogue with grounded document:} Second, we recruit annotators to generate multi-turn conversations that are related to a news article. In this step, the significant difference between our work and others is that we have few restrictions or assumptions for participants. For example, we do not have an explicit goal during conversation as proposed in \cite{wu-etal-2019-proactive} that requires the conversation should transfer to a different entity within the give topic. In addition, two participants in our data collection both have access to the news article rather than that only one participant has access to the material as an expert and the other does not as an apprentice, as proposed in \cite{dinan2019wizard}. During our conversation, we only have the following three very limited requirements:

\begin{itemize}
\item The length of dialogue must be no less than 10 turns for each participant and the content of the news article must be mentioned. But, we do not require how much of the content needs to be involved and how the content should be mentioned. Therefore, once anything in the news is touched, the participants can shift the topic immediately as long as the shift is smooth. The shift can be anything such as content related to the topic but unrelated to the news article, or even unrelated to the topic. Table 1 and Table 2 in the supplemental material give a complete example. News from Table 1 is the common topic for the dialogue in Table 2. It is interesting to find the topic can be shifted to a Chinese TV show ``Where Are We Going? Dad'' from the news about German F1 racing driver Michael Schumacher.
\item Every conversation must happen in a scenario. The participants can decide the scenario in their preference or they can choose a scenario that can easily trigger the initial topic. Table 3 and Table 4 in the supplemental material give such an example. The news is from technology category and is about a newly released wireless headset from Google. The dialogue happens on a playground where lots of people are doing exercise.
\item Once the participants read the above two instructions, they can talk about anything they want as long as the conversation goes as naturally as possible and follows the human logic. One example is from Table 5 and Table 6 of the supplemental material. The news is about a newly released electronic game. But we find few utterances are about the game itself. Two participants talk a lot about their experience of playing the game in childhood and plan to play it together in the near future. At the end of the dialogue, the participants even exchange the account IDs for playing another game.
\end{itemize}

We employ a data supplier in China to carry out the dialogue collection project. We cooperate closely with our partner, monitoring every detail in the process. We enforce strict guidelines to control the quality, checking if there is repetition between each utterance and the news passage, making sure the participant combinations are as diverse as possible, monitoring utterances length to eliminate any perfunctory behavior, etc. We also sample the dialogues and manually read them as one of our quality management methods. Any dialogue that fails our test would be returned and rewritten until it passes our examination. We pay our data supplier roughly $\$50,000$ for the task.

In our data collection, we do not require our supplier to provide fine-grained sentence-level annotation for linguistic features due to the following observations: 1) the dialogue pattern in our corpus is largely oral and very flexible for accurate and efficient annotations; 2) there is no obvious correspondence between the sentences in the document and the utterances in the dialogue. However, extra annotations from other parties in the community are always welcome.

\subsection{Corpus Statistics}
Table~\ref{corpus_sta} summarizes the regular information about NaturalConv. In the following, we describe two specific metrics on related corpora and ours to present our corpus's features.

\begin{table*}[ht]
\begin{center}
\begin{tabular}{lccccccc} \hline
 &  \textbf{Sports} & \tabincell{c}{ \textbf{Ent}} & \tabincell{c}{ \textbf{Tech}}&\textbf{Games} &  \textbf{Edu} &  \textbf{Health} & \ \textbf{Total} \\ \hline
\textbf{\# document} & 3,124   & 1,331& 1,476 & 103 & 414  & 52 & 6,500\\
\textbf{\# dialogues} & 9,740   & 4,403& 4,061 & 308 & 1,265  & 142 & 19,919\\
\textbf{\# dialogues per document} & 3.1   & 3.3& 2.8 & 3.0 & 3.1  & 2.7 & 3.0\\
\textbf{\# utterances} & 195,643   & 88,457 & 81,587 & 6,180 & 25,376& 2,852 &400,095\\
\textbf{Avg. \# utterances per dialogue}   & 20.1 & 20.1 &  20.1 & 20.1 & 20.1& 20.1&20.1 \\

\textbf{Avg. \# tokens per utterance}   & 12.0 & 12.4 & 12.3 & 12.1& 12.6& 12.5 &12.2\\
\textbf{Avg. \# characters per utterance}   & 17.8 & 18.1 & 18.6 & 17.8& 18.1&18.3&18.1 \\
\textbf{Avg. \# tokens per dialogue}   & 241.1 & 248.2 &  247.5 & 242.9 & 248.3& 251.1 &244.8\\
\textbf{Avg. \# characters per dialogue}  & 357.5 & 363.2 &  372.8 & 356.5 & 356.5& 368.0 &363.1\\ \hline

\end{tabular}
\caption{Statistic of NaturalConv.}
\label{corpus_sta}
\end{center}
\end{table*}

\textbf{Similarity between document and dialogues}: As aforementioned, our conversations include indirect content in respect with the document. We use the BLEU score similarity between document and dialogue to measure in our dataset and other existing datasets how much content people talk about are directly from the background document. A lower similarity measure indicates the dialogue is less the repetition of the document and potentially more natural and informative. In this evaluation, we conduct the comparison between our dataset, CMU DoG \cite{zhou-etal-2018-dataset}, DuConv \cite{wu-etal-2019-proactive} and KdConv \cite{zhou2020kdconv}. In these datasets, each dialogue has a dialogue-level grounding information. CMU DoG and ours are plain text and the other two are structured KG. Obviously, from Table~\ref{corpus_sim}, we find our dataset has the lowest BLUE1 score and significantly lower BLUE2 score compared to other datasets.

\begin{table}[ht]
\begin{center}
\begin{tabular}{ccc} \hline
 &  \textbf{BLEU1} & \textbf{BLEU2} \\ \hline
\textbf{CMU DoG} & 23.7   & 9.64  \\
\textbf{DuConv} & 16.53   & 10.58 \\
\textbf{KdConv} & 35.69   & 26.27\\
\textbf{NaturalConv} & 16.17   & 6.13  \\
\hline
\end{tabular}
\caption{Statistic of similarity between grounding document and dialogue among different dataset.}
\label{corpus_sim}
\end{center}
\end{table}

\textbf{Variability between dialogues}: Since we have few restrictions for annotators, they are allowed not only to talk on different aspects within the document, but also to shift topics easily. We believe this will lead to better variability of dialogues given the same or similar documents. To prove our point, we measure the variability between dialogues from different pairs of annotators when given the same background document/knowledge grounding. Specifically, for each document/knowledge in CMU DoG, DuConv, KdConv, and NaturalConv, we randomly choose 3 corresponding dialogues respectively, calculate the BLEU scores for each possible pair of the 3 dialogues, and average them. Finally, we average the scores across all the documents/knowledge to represent the overall variability of the corpus. The higher of score means the lower variability between dialogues. Results in Table~\ref{corpus_sim2} indicate our dataset has the best variability.

\textbf{Human evaluation on naturalness}: We perform manual evaluation of dialogue naturalness for DuConv, KdConv, and NaturalConv. We randomly selected 100 sessions of dialogues from each of the three corpora. Dialogue-level evaluations are performed by two annotators with three grades: natural (3), fair (2), unnatural (1). Evaluation shows our corpus has the best averaged naturalness score of 2.8. DuConv and KdConv have scores of 2.4 and 2.0 respectively.

\begin{table}[ht]
\begin{center}
\begin{tabular}{ccc} \hline
 &  \textbf{Avg-BLEU1} & \textbf{Avg-BLEU2} \\ \hline
\textbf{CMU DoG} & 33.15   & 14.62  \\
\textbf{NaturalConv} & 32.36   & 12.56  \\
\hline
\end{tabular}
\caption{Statistic of similarity between dialogues under same topic among different dataset.}
\label{corpus_sim2}
\end{center}
\end{table}

\section{Methods}

In this section, we discuss the methods we use for conversation modeling and response generation with the collected NaturalConv corpus. Both the retrieval-based and generation-based methods are evaluated. To further explore the role of the document grounding on dialogue, we extend the generation models to integrate the retrieved document contents most related to the current dialogue context through attention mechanism.

\subsection{Retrieval-based Model}
Given a dialogue context $\mathbf{X}$, the retrieval-based dialogue system responds to the context via searching for the best response $\mathbf{y}$ from the NatrualConv corpus. We adopt an IR model by finding the most similar query in the retrieval corpus and then utilizing its response as the result. Similarity is measured by the BM25 index between bags of words.

Recently, the BERT based retrieval dialogue models~\cite{whang2019domain} have shown promising performances in dialogue systems. We further incorporate the BERT model to re-rank outputs of the BM25 retrieval model. This Retrieval-BERT model is fine-tuned from the ``bert-base-chinese'' backbone for sequence classification on the training data with both the ground truth response labeled as ``1'' and top $K-1$ responses from the BM25 retrieval method labeled as ``0''. During inference, it re-ranks the top $K$ response from the BM25 retrieval method given the dialogue context as query according to the sequence classification scores of the fine-tuned model.

\subsection{Generation-based Model}

In our multi-turn conversation setting, the generation-based dialogue models take the concatenation of the past $k$ dialogue utterances as input $\mathbf{X}=\{\mathbf{x}_1, \mathbf{x}_2,...,\mathbf{x}_k\}$, and outputs a natural-language response consisting of a sequence of words $\mathbf{y}=\{y_1, y_2, ..., y_n\}$, where $n$ is the maximum possible number of words in the response sequence. The training of generation-based dialogue models requires a training dataset $\mathcal{D}=\{(\mathbf{X}^i, \mathbf{y}^i)_{i=1}^N\}$ containing $N$ gold input-output dialogue pairs $(\mathbf{X}^i,\mathbf{y}^i)$. To train parameter $\theta$ of the generative model, we use Maximum Likelihood Estimation (MLE) to minimize the loss $\mathcal{L}=\sum_{i=1}^N \mathcal{L}^i(\mathbf{X}^i, \mathbf{y}^i; \theta)$, where
\[ \mathcal{L}^i(\mathbf{X}^i, \mathbf{y}^i; \theta) = - \sum_{t=1}^{|\mathbf{y}^i|} \log P(y_t^i|\mathbf{X}^i, y_{<t}^i |\theta). \]

We implement the generation-based model as the Seq2Seq model consisting of an encoder and a decoder. Different encoder-decoder structures including GRU and LSTM with attention mechanism, as well as the Transformer encoder-decoder model are used.

\subsection{Model with Document Grounding}

To further incorporate the document grounding information in the generation-based models, we split the document $\mathbf{S}$ into a sequence of sentences $\mathbf{S}=\{\mathbf{s}_1, \mathbf{s}_2, ..., \mathbf{s}_m\}$. Given the dialogue context input $\mathbf{X}=\{\mathbf{x}_1, \mathbf{x}_2,...,\mathbf{x}_k\}$ to the generation model consisting of $k$ contextual dialogue utterances, we retrieve from $\mathbf{S}$ the sentence $\mathbf{s}_*$ that is most similar to the most recent dialogue utterance $\mathbf{x}_k$. The generation model would then take the concatenated $(\mathbf{s}_*, \mathbf{X})$ as input to generate the response $\mathbf{y}$.

To train the model with document grounding, we minimize the loss $\mathcal{L}=\sum_{i=1}^N \mathcal{L}^i(\mathbf{X}^i, \mathbf{y}^i, \mathbf{s}_*^i; \theta)$, where
\[ \mathcal{L}^i(\mathbf{X}^i, \mathbf{y}^i, \mathbf{s}_*^i; \theta) = - \sum_{t=1}^{|\mathbf{y}^i|} \log P(y_t^i|\mathbf{X}^i, \mathbf{s}_*^i, y_{<t}^i |\theta). \]

We implement the model with document grounding also as the Seq2Seq model with the encoder-decoder structure. We use the attention mechanism for both GRU and LSTM models to ensure the document information $\mathbf{s}_*$ can be incoproated. The Transformer encoder can incorporate $\mathbf{s}_*$ through its self-attention mechanism.  We denote these generation models incorporating docs as ``GRU with Doc'', ``LSTM with Doc'', and ``Transformer with Doc''.

\section{Experiments}

We conduct experiments to provide benchmark results for the NaturalConv dataset. Both the results of retrieval-based models and generation-based are evaluated. Furthermore, we evaluate the performance of models with document grounding, and provide discussions on the results.

\subsection{Implementation Details}

We implement LSTM, GRU, BERT, and Transformer models with PyTorch. The experiments are performed on Nvidia Tesla P40 GPUs. The LTP~\cite{che2010ltp} Chinese word segmentation tool is used for tokenization.

Our Retrieval model uses BM25 index to retrieve the most related response in the corpus. The Retrieval-BERT model re-ranks the top $K=10$ retrieved responses. Our GRU network consists of the one-layer bi-directional GRU encoder and the one-layer GRU decoder. Its embedding size is set to 300, and the hidden state size is set to 800. The LSTM network consists of a two-layer bi-directional LSTM encoder and and a two-layer LSTM decoder. Both the embedding size and the hidden state size of the LSTM model are set to 500. The Transformer model contains a six-layer encoder and a six-layer decoder, with the embedding size, hidden unit size, and attention head number to be 1024, 4096, and 16, respectively.

ADAM is used to optimize the GRU, LSTM and Transformer models,  with the initial learning rate set to be $5 \times 10^{-5}$ for GRU, $1 \times 10^{-3}$ for LSTM, and $5 \times 10^{-4}$ for Transformer, respectively.

\subsection{Metrics}

Our proposed models are tested with the following automatic evaluation metrics: 1) BLEU-1 and BLEU-2 scores~\cite{tomeh2009complexity}, 2) F1 score~\cite{wu-etal-2019-proactive}, 3) DISTINCT-1 and DISTINCT-2 scores~\cite{li2015diversity}, 4) BERTScore~\cite{bert-score}.

The BLEU-1/2 scores evaluate the token (word) level similarity between the output response and the reference response. The F1 score, comparatively, evaluates the Chinese character level similarity between the output response and the reference response. We further use the DISTINCT-1/2 scores to evaluate the diversity of the generated sentences. Finally, the recently proposed BERTScore is used to obtain a similarity measure that does not require exact match of tokens or characters. We denote the mean values of the BERTScore precision, recall, and F1 measure over all testing pairs as $P_\textrm{\scriptsize{BERT}}$, $R_\textrm{\scriptsize{BERT}}$, and $F_\textrm{\scriptsize{BERT}}$, respectively. Our backbone model for BERTScore evaluation is the ``bert-base-chinese'' BERT model released by \cite{devlin2018bert}.

\subsection{Data Split}

We split different documents and their corresponding dialogues from the NaturalConv corpus into the train, dev, and test sets, respectively. As a result, different dialogues belonging to the same grounding documents can only appear simultaneously either in train, dev, or test set. The total number of documents in different topics, as well as the total number of dialogue pairs for each set are presented in Table~\ref{exp:split}. The data split will be released together with the corpus.

\begin{table}[ht]
\begin{center}
\begin{tabular}{c|ccc}
\hline
 &  Train & Dev & Test \\
\hline
\# Doc - Sports &9500  & 120 & 120\\
\# Doc - Ent & 4283 & 60 & 60\\
\# Doc - Tech & 3941 & 60 & 60\\
\# Doc - Games & 292 & 8 & 8\\
\# Doc - Edu & 1233 & 16 & 16\\
\# Doc - Health & 126 & 8 & 8\\
\hline
\# Dialogue Pairs & 369802 & 5183  & 5191\\
\hline
\end{tabular}
\caption{Statistics of our train, dev, and test sets.}
\label{exp:split}
\end{center}
\end{table}

\subsection{Results}

The results for generation-based conversation models are given in Table~\ref{exp:baseline}. We further provide the performance of models incorporating document grounding in Table~\ref{exp:grounding}.

\begin{table*}[ht]
\begin{center}
\begin{tabular}{c|c|c|c|c}
\hline
 & BLEU-1/2 & DISTINCT-1/2 & F1 & $P_\textrm{\scriptsize{BERT}}$ / $R_\textrm{\scriptsize{BERT}}$ / $F_\textrm{\scriptsize{BERT}}$ \\
\hline
\textbf{Retrieval} & 23.30 / 13.12 & \textbf{8.48} / \textbf{43.21} & 23.39 & 63.78 / 64.22 / 63.90 \\
\textbf{Retrieval-BERT} & 24.96 / 13.82 & 8.27 / 42.31 & 24.87 & 65.35 / 64.87 / 65.01\\
\hline
\textbf{GRU} & \textbf{27.89} / \textbf{14.23} & 1.80 / 8.17 & 26.61 & 67.49 / \textbf{65.35} / \textbf{66.32} \\
\textbf{LSTM} & 26.09 / 13.35 & 0.98 / 4.30 & \textbf{26.65} & \textbf{67.97} / 64.49 / 66.09 \\
\textbf{Transformer} & 25.17 / 12.39 & 2.91 / 15.32 & 25.73 & 65.37 / 64.55 / 64.84 \\
\hline
\end{tabular}
\caption{Performances of the retrieval-based and generation-based dialogue models without incorporating information from the document on the NaturalConv corpus.}
\label{exp:baseline}
\end{center}
\end{table*}

\begin{table*}[ht]
\begin{center}
\begin{tabular}{c|c|c|c|c|c}
\hline
 & BLEU-1/2 & DISTINCT-1/2 & F1 & $P_\textrm{\scriptsize{BERT}}$ / $R_\textrm{\scriptsize{BERT}}$ / $F_\textrm{\scriptsize{BERT}}$ & Manual (1-3) \\
\hline
\textbf{GRU} & \textbf{27.89} / 14.23 & 1.80 / 8.17 & 26.61 & 67.49 / 65.35  / 66.32 & 1.97\\
\textbf{GRU with Doc} & 27.86 / 14.24 & 1.87 / 8.73 & 26.70 & 67.39 / 65.32 / 66.25 & 2.03\\
\hline
\textbf{LSTM} & 26.09 / 13.35 & 0.98 / 4.30 & 26.65 & 67.97 / 64.49 / 66.09 & 2.10 \\
\textbf{LSTM with Doc} & 26.79 / \textbf{14.54}  & 2.13 / 9.49 & \textbf{28.08} & \textbf{68.50} / \textbf{65.60} / \textbf{66.92} & \textbf{2.16} \\
\hline
\textbf{Transformer} & 25.17 / 12.39 & \textbf{2.91} / \textbf{15.32} & 25.73 & 65.37 / 64.55 / 64.84 & 2.04 \\
\textbf{Transformer with Doc} & 24.47 / 13.12 & 2.77 / 14.35 & 27.01 & 67.39 / 65.06 / 66.08 & 2.09 \\
\hline
\end{tabular}
\caption{Performances of the generation-based dialogue models without or with incorporating information from the document on the NaturalConv corpus.}
\label{exp:grounding}
\end{center}
\end{table*}

\textbf{Comparison between dialogue models.} From Table~\ref{exp:baseline}, we can see the GRU and LSTM models in many cases outperform the retrieval-based model in terms of different similarity metrics including BLEU-1/2, F1, and BERTScore. On the other hand, the retrieval-based model significantly outperforms the generation-based models in the DISTINCT-1/2 metrics. It indicates the GRU and LSTMs models can generate dialogue responses that are more similar to the golden responses. However, these generation-based Seq2Seq models are still not capable enough to generate dialogue responses that are as diverse as the human responses that are retrieved by the retrieval model.

\textbf{Performances of generation-based models.} We can further compare the performances between different generation-based dialogue models in Table~\ref{exp:baseline}. In our experiment, GRU, LSTM, Transformer are all trained only with the NaturalConv corpus. GRU and LSTM in general have similar performances in terms of the similarity measures between their generated responses and the ground truth responses. Comparatively, the Transformer model, which is significantly bigger than GRU and LSTM in terms of model size, performs slightly worse than both GRU and LSTM in similarity measures including Bleu-1/2, F1, and BERTScore on our NaturalConv corpus with 369,802 dialogue pairs for training. On the other hand, the Transformer model obviously outperforms both GRU and LSTM in DISTINCT-1/2 measures, indicating its responses are more diverse than those of GRU and LSTM.

\textbf{Performances with document grounding.} In Table~\ref{exp:grounding}, we compare the performances of GRU, LSTM, and Transformer models without or with incorporating information from the document through our model with document grounding. In this experiment, we can observe that the performances of GRU and GRU with Doc are similar in all the metrics. The LSTM with Doc model improves the LSTM model in similarity measures including BLEU-1/2, F1, and BERTScore, as well as in diversity measures DISTINCT-1/2. Similar improvements can be found in the Transformer with Doc model in comparing to the baseline Transformer.

\textbf{Human evaluation of generation-based models.} We perform manual generation quality evaluation with the randomly selected 100 queries in the test set for each model in Table~\ref{exp:grounding}. The responses are evaluated by two annotators with the overall quality score in three grades: good (3), fair (2), bad (1). The averaged scores in Table~\ref{exp:grounding} show slight performance gains for models with document comparing to their corresponding models without document.

In summary, the improvements from incorporating the document information for dialogue response generation is still not significant. It indicates that the current methodology could still be limited in discovering/exploiting from the document the information that is more likely to be used by human beings in the multi-turn topic-driven conversation setting. Moreover, considering NaturalConv dialogue includes information outside of the document, utilizing knowledge outside of the document or corpus could also be beneficial to further improve the performance.

\section{Conclusion}
In this paper, we propose a Chinese multi-turn topic-driven conversation generation corpus, NaturalConv. It contains 400K utterances and 19.9K dialogues, with an average number of 20.1 turns. 
Each dialogue is based on a shared topic and two participants are free to talk anything as long as any one specific aspect from the topic is mentioned. The participants are also required to assume a scenario for the conversation. Therefore, the dialogue contains various conversation elements such as chitchat, discussions about the topic, any possible extensions of the topic, etc. We believe this dataset provides a good benchmark to evaluate the ability to model topic-driven free-style conversations. In addition, we provide results of several benchmark models to facilitate further research. Experiments demonstrate that our current models can not provide significant improvement by introducing document knowledge, therefore there is much room in topic-grounded conversation modeling for future work.

\section*{Acknowledgements}
We thank all the annotators for annotating the dialogues. The views and opinions expressed in the dataset including the documents and the dialogues do not necessarily reflect those of Tencent or the authors of this paper.

\bibliography{ref}

\end{CJK*}
\end{document}